\documentclass[twoside,11pt]{article}

\usepackage{automl2019}
\usepackage{nicefrac}
\usepackage{amsmath}
\usepackage{amsfonts}
\usepackage{amssymb}
\usepackage{bbm}
\usepackage{booktabs}       % professional-quality tables
\usepackage{color}
\usepackage{graphicx}
\usepackage{listings}
\usepackage{subcaption} 
\graphicspath{ {./images/} }

\usepackage{color}
 
\definecolor{codegreen}{rgb}{0,0.6,0}
\definecolor{codegray}{rgb}{0.5,0.5,0.5}
\definecolor{codepurple}{rgb}{0.58,0,0.82}
\definecolor{backcolour}{rgb}{0.95,0.95,0.92}
 
\lstdefinestyle{mystyle}{
    backgroundcolor=\color{backcolour},   
    commentstyle=\color{codegreen},
    keywordstyle=\color{magenta},
    numberstyle=\tiny\color{codegray},
    stringstyle=\color{codepurple},
    basicstyle=\footnotesize,
    breakatwhitespace=false,         
    breaklines=true,                 
    captionpos=b,                    
    keepspaces=true,                 
    numbers=left,                    
    numbersep=5pt,                  
    showspaces=false,                
    showstringspaces=false,
    showtabs=false,                  
    tabsize=2
}

\newcommand{\dataset}{{\cal D}}

\newcommand\E{\mathbb E}

\jmlrheading{Charles Weill et al.}

\ShortHeadings{AdaNet: A Scalable and Flexible Framework for Automatically Learning Ensembles}{Weill et al.}
\firstpageno{1}

\begin{document}

\title{AdaNet: A Scalable and Flexible Framework for Automatically Learning Ensembles}

\author{\name Charles Weill$^1$, Javier Gonzalvo$^1$, Vitaly Kuznetsov$^1$,
    Scott Yang$^3$, Scott Yak$^1$, Hanna Mazzawi$^1$, Eugen Hotaj$^1$,
    Ghassen Jerfel$^{1,2}$, Vladimir Macko$^1$, Ben Adlam$^1$, Mehryar Mohri$^{1,3}$, \and Corinna Cortes$^1$ \\
       \addr  $^1$Google AI*, New York, New York, United States; $^2$Duke University, Durham, North Carolina, United States; $^3$Courant Institute of Mathematical Sciences, New York, New York, United States  \\
      *Corresponding authors: \email \{weill, xavigonzalvo\}@google.com}

\maketitle

\begin{abstract}%   <- trailing '%' for backward compatibility of .sty file
% AdaNet is a lightweight TensorFlow-based framework for automatically learning high-quality models with minimal expert intervention. It is designed to be a general AutoML framework that is flexible and scalable while providing learning guarantees. Importantly, AdaNet provides a general framework not only learning a neural network architecture, but also for learning to ensemble to obtain even better models. This project is based on the AdaNet algorithm, presented in ``AdaNet: Adaptive Structural Learning of Artificial Neural Networks'' \cite{cortes17adanet}, for learning the structure of a neural network as an ensemble of subnetworks. The framework's API allows users to define model search spaces and distribute the search over a clusters of CPUs, GPUs or TPUs. The code is open-source and available at \url{https://github.com/tensorflow/adanet}.
% AdaNet is a lightweight TensorFlow-based framework for automatically learning high-quality models with minimal expert intervention. AdaNet builds on recent AutoML efforts to be fast and flexible while providing learning guarantees. Importantly, AdaNet provides a general framework for not only learning a neural network architecture, but also for learning to ensemble to obtain even better models. This project is based on the AdaNet algorithm, presented in “AdaNet: Adaptive Structural Learning of Artificial Neural Networks” at ICML 2017, for learning the structure of a neural network as an ensemble of subnetworks. The code is open-source and available at https://github.com/tensorflow/adanet
AdaNet is a lightweight TensorFlow-based \citep{tensorflow2015-whitepaper} framework for automatically learning high-quality ensembles with minimal expert intervention. Our framework is inspired by the \textit{AdaNet} algorithm \citep{cortes17adanet} which learns the structure of a neural network as an ensemble of subnetworks. We designed it to: (1) integrate with the existing TensorFlow ecosystem, (2) offer sensible default search spaces to perform well on novel datasets, (3) present a flexible API to utilize expert information when available, and (4) efficiently accelerate training with distributed CPU, GPU, and TPU hardware. The code is open-source and available at: \url{https://github.com/tensorflow/adanet}.
\end{abstract}

\section{Introduction}
Recent years have seen the successful application of machine learning to a wide range of real-world problems. However each success requires making countless expert decisions in all aspects of data collection, feature engineering, and model search \citep{mendoza-automlbook18a}. The \textit{AutoML} field has the objective to automate parts of this pipeline to produce high-quality models, free up expert time, and make machine learning more accessible. While the field has existed for many year, AutoML with large datasets and complex models is recently a viable option thanks to the expansion in computational power available through specialized hardware and cloud compute services.

% Ensemble learning, the art of combining different machine learning (ML) model predictions, is widely used with neural networks and in AutoML frameworks to achieve state-of-the-art performance, benefiting from a rich history and theoretical guarantees to enable success at challenges such as the Netflix Prize and various Kaggle competitions.

Ensemble methods, one particular family of modeling techniques, consistently achieve state-of-the-art performance at challenges such as the Netflix Prize and Kaggle competitions \citep{zhou2012ensemble}. They benefits from a rich history of theoretical guarantees \citep{DBLP:journals/jcss/FreundS97}, as well as new studies on their generalization capabilities \citep{cortes14deepboosting}. However, building high-quality ensembles requires significant expertise, such as choosing the right base models, and knowing how to train them and combine their outputs. 
%  Several open-source AutoML frameworks such as Auto-Weka by \cite{thornton2013autoweka} and AutoKeras by \cite{jin@autokeras} tackle this problem by searching over a space of models. 
In this paper, we introduce AdaNet, our scalable and flexible TensorFlow framework for automatically learning ensembles.

\section{Related Work}

When designing AdaNet for our research and production needs within Google, we gave ourselves the following constraints. On one hand, we wanted a framework that could automatically produce a high-quality model given an arbitrary set of features and a model search space. On the other hand, we wanted it to build ensembles from productionized TensorFlow models to reduce churn, reuse domain knowledge, and conform with business and explainability requirements. Our framework should handle datasets containing thousands to billions of examples. Finally, we needed it to utilize available distributed compute and leverage accelerators when available, because training a single production model may take several days on hundreds of machines.

Several open-source AutoML frameworks, such as auto-sklearn \citep{feurer2015autosklearn} and auto-pytorch \citep{mendoza-automlbook18a}, encode the expertise necessary to automate ensemble construction, and achieve impressive results. In comparison, our framework is built using TensorFlow to facilitate integration with TensorFlow-based production infrastructure and tooling. Furthermore, it is uniquely designed to efficiently execute on hundreds of heterogeneous workers in a distributed cluster and run on TPU. We designed AdaNet to meets our aforementioned needs, and have open sourced it to share with the entire AutoML community across companies and universities to accelerate open research.

% The framework presented in this paper is a scalable TensorFlow implementation of the \textsc{AdaNet} algorithm by \cite{cortes17adanet}, which generalizes the algorithm to other settings (see Figure~\ref{fig:search}). We present three novel improvements: (1) flexible addition and pruning of the ensemble; (2) relaxed constraint on subnetwork type (e.g., subnetworks can now be convolutional, recurrent, etc); and (3) extended support for complexity measures other than Rademacher bounds. 
 We structured this paper as follows. Section~\ref{sec:overview} presents an overview of the system design. In Section~\ref{sec:implementation} we outline framework implementation details and introduce our novel \textit{adaptive computation graph} for handling TensorFlow limitations. Finally, in Section~\ref{sec:results} we present some applications of AdaNet in production at Google.

\section{Overview}
\label{sec:overview}

An AdaNet run produces an \textit{ensemble} model $f$ composed of $k$ \textit{subnetworks} $h_i$ (a.k.a \textit{base learners} or \textit{weak learners} in the literature \citep{zhou2012ensemble}) where $ i \in [0,k-1], k \geq 1$. These ensembles are model-agnostic: subnetworks can be as simple as an if-statement, or as complex as convolutional or recurrent neural networks. 

% The only requirement is that for a given input tensor $x$, the outputs of each subnetwork $h_i(x)$ are combined into a single prediction $f(x)$.

% In our framework, every model is an \textit{ensemble} composed of one or more \textit{subnetworks} whose outputs are combined via an \textit{ensembler}. The emphasis on solving the ensemble learning problem first carries several practical benefits. Ensembles are model-agnostic, so a subnetwork can be as complex as a deep neural network, or as simple as an if-statement; the only requirement is that for a given input tensor $x$, subnetwork outputs $h_i(x)$ can be combined by the ensembler to output a single prediction $f(x)$. Furthermore, the design of a generic framework for producing a high-quality ensemble must also support AutoML features like architecture search and hyperparameter tuning. Finally, ensembles have favorable theoretical properties, which we discuss in Section~\ref{sec:learning_guarantees}.

% \subsection{Adaptive search for an optimal ensemble}

\begin{figure}  
\begin{center}  
\begin{subfigure}[b]{0.6\textwidth}  
\includegraphics[width=9cm,keepaspectratio]{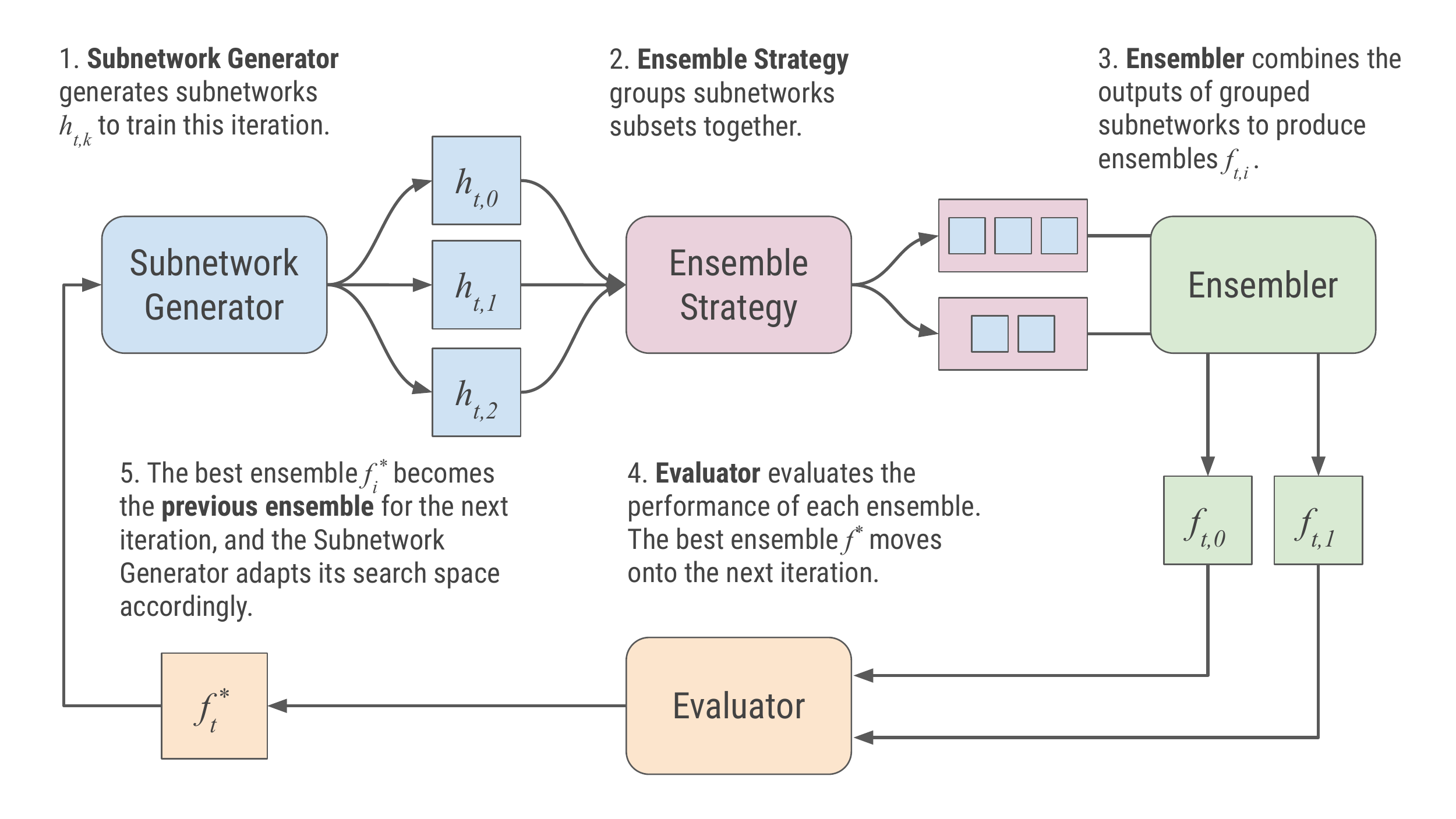}
% \label{fig:lifecycle}
\end{subfigure} 
\begin{subfigure}[b]{0.3\textwidth}
\includegraphics[width=4.5cm,keepaspectratio]{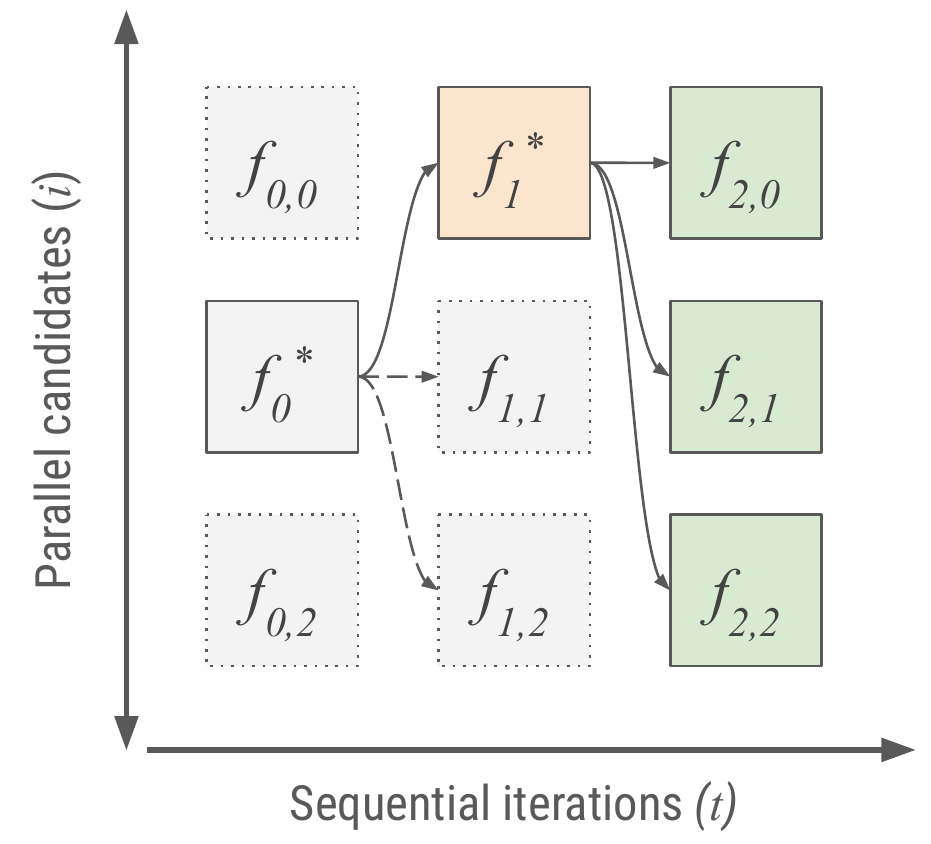}
% \label{fig:search}
\end{subfigure} 
\end{center}
\vspace{-0.2in}
\caption{\small \sl Overview of the AdaNet adaptive search strategy.}
\label{lifecycle}
\end{figure}  

AdaNet combines two orthogonal ensembling paradigms \citep{zhou2012ensemble}: parallel (similar to bagging) and sequential (similar to boosting). Together, these form the axes of the \textit{adaptive search space} that the framework iteratively explores for an optimal ensemble (Figure~\ref{lifecycle}). The search space is defined by the combination of \textit{Subnetwork Generators} which generate candidate subnetworks $h_{t}$ for iteration $t$, \textit{Ensemble Strategies} which form discrete groups of subnetworks, and \textit{Ensemblers} which combine the predictions of grouped subnetwork into ensembles $f_{t}$. The framework is responsible for managing and training these ensembles and subnetworks. The \textit{Evaluator} evaluates the candidate ensembles once they are finished training to select and fix $f_{t}^{*}$, the ensemble with the best performance according to the objective, along with its component subnetworks. The search then proceeds to  iteration $t+1$, where the \textit{Subnetwork Generator} adapts its search space according to the previous best ensemble. For example, if the subnetwork search space explores increasingly deeper neural networks, and the deepest subnetwork in the ensemble is $l$ layers deep, the Subnetwork Generator could generate one candidate subnetwork with $l$ hidden layers, and another with $l+1$.

\section{Implementation}
\label{sec:implementation}

In this section we cover the implementation details of the AdaNet framework including APIs, its novel \textit{adaptive computation graph}, and distributed training strategies.

\subsection{Application Programming Interface}

AdaNet extends \texttt{tf.estimator.Estimator} by \cite{cheng2017estimator} to encapsulate training, evaluation, prediction and export for serving. This abstraction enables the same user code to run on different hardware including CPUs, GPUs and TPUs. To specify the target task, such as regression, classification, or multi-task, the user passes a \texttt{tf.estimator.Head} instance when constructing the \texttt{adanet.Estimator}. Being an \texttt{Estimator} allows AdaNet models to be drop-in replacements for existing Estimator models, and integrate with tools in the TensorFlow open-source ecosystem (\url{https://github.com/tensorflow}) like TensorFlow Hub, Model Analysis, and Serving. A Keras \citep{chollet2015keras} API equivalent is under development. 

\lstset{language=Python, stringstyle=\ttfamily\color{blue}}
\begin{lstlisting}
import adanet
import tensorflow as tf
estimator = adanet.AutoEnsembleEstimator(
    head=tf.contrib.estimator.multi_class_head(n_classes=10),
    ensemblers=[adanet.ensemble.ComplexityRegularizedEnsemble()],
    ensemble_strategies=[adanet.ensemble.GrowStrategy()],
    candidate_pool={
        "linear": tf.estimator.LinearCLassifier(...),
        "dnn":    tf.estimator.DNNClassifier(...),
        "gbdt":   tf.estimator.BoostedTreesClassifier(...)})
estimator.train(...).evaluate(...).export_saved_model(...)
\end{lstlisting}

Users specify their search spaces using the \texttt{adanet.subnetwork} package to define how subnetworks adapt at each iteration, and the \texttt{adanet.ensemble} package to define how ensembles are composed, pruned, and combined. AdaNet provides \texttt{AutoEnsembleEstimator} to users who want a higher-level API for defining a search space in only a few lines of code using canned \texttt{Estimators} like \texttt{DNNEstimator} and \texttt{BoostedTreesClassifier}. 

For visualizing model performance during training, the framework integrates with TensorBoard \citep{wongsuphasawat2017tensorboard}. When training is finished, the framework exports a TensorFlow SavedModel that can be deployed with TensorFlow Serving or similar services.

\subsection{Adaptive Computation Graph}

\begin{figure}  
\begin{center}  
\includegraphics[width=14cm,keepaspectratio]{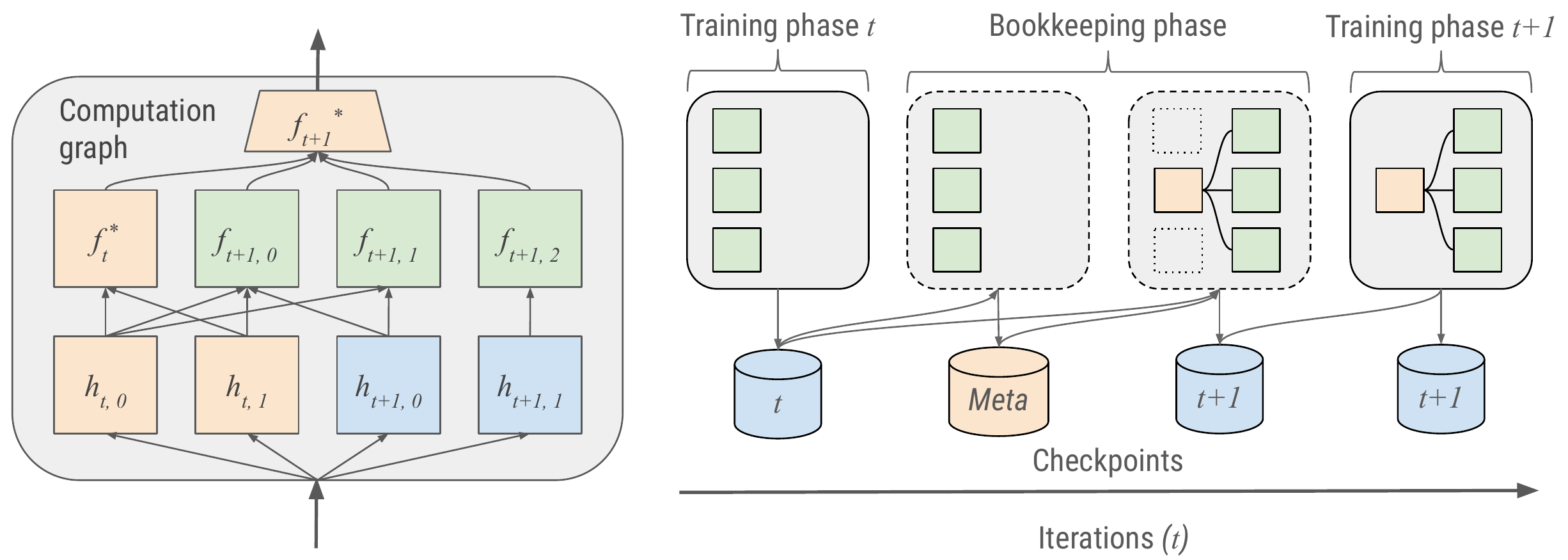}
\end{center}
\vspace{-0.2in}
\caption{\small \sl The adaptive computation graph. Within an iteration, a static computation graph contains multiple subnetworks and ensembles, including the best ensemble and subnetworks from the previous iteration. Each new ensemble candidate is composed of a subset of the present subnetworks. The iteration tracks and exposes the predictions of the best ensemble at each training step. Across iterations, AdaNet implements an adapts its computation graph for the next iteration.}
\label{single_graph}
\end{figure}

A common mechanism for training an ensemble is to first train each base learner in a separate process, and then ensemble them in a second phase \citep{Caruana:2004:ESL:1015330.1015432}. However, accelerators such as TPUs do not allow multiple processes to share the same hardware, which limits the number of candidates that can be trained in parallel. Instead, AdaNet creates all candidate within the same computation graph and session, including new subnetworks and ensembles for iteration $t$, and the best ensemble  and  corresponding subnetworks from $t-1$. (see Figure~\ref{single_graph}). This design allows candidates to share tensors. For example, subnetworks can share the same input pipeline and  ensembles can share subnetwork outputs. It also makes many complex ensembling routines possible such as knowledge distillation \citep{hinton2015distilling} and Born Again Networks \citep{furlanello2018born}, and can straightforwardly be extended to more complicated ones like population based training \citep{jaderberg2017pbt}. Furthermore, having all the candidates within the same graph allows compilers such as the TensorFlow compiler and XLA to optimally place ops on logical devices, which is particularly important for maximizing multi-core accelerator utilization such as when training many small subnetworks on a TPU.

Across iterations, the computation graph must evolve. However, TensorFlow and XLA were designed for a static computation graph, which limits our ability to create new operations during training. There are a number of workarounds: one is to dynamically store a model in a resource variable. This doesn't work for us, since it would require users to rewrite their models in low-level operations in C++ instead of Python, which would drastically increase the cost of adopting AdaNet and would limit flexibility. Another is to implement the framework using Eager execution, which unfortunately supports neither distributed nor TPU training.

The solution is to modify the train loop to create an \textit{adaptive computation graph} as described in Figure~\ref{single_graph}: after completing training of all candidates in iteration $t$, AdaNet begins a \textit{bookkeeping} phase. During this phase it reconstructs the graph with the evaluation dataset and evaluates all the candidates to determine the best ensemble $f_t^*$ for iteration $t$. Next it serializes metadata about the architecture of $f_t^*$, and uses this metadata to construct a new graph for $t+1$. The new graph includes $f_t^*$ and all the new subnetworks $h_{t+1}$ and ensembles $f_{t+1}$, and warm-starts the variables of $f_t^*$ from the most recent checkpoint. Finally, it creates a new checkpoint with the modified graph for $t+1$ and increments the \textit{iteration number} variable in the new checkpoint. When Estimator resumes training, it will construct the new graph based on the architecture metadata from iteration $t$, and will restore variables from the new checkpoint, and treat it as a static graph. Evaluate and predict have no effect on the iteration number so their methods require no modification.

\subsection{Distributed Training}

When training on small datasets or when debugging, AdaNet can be executed in single process. To speed up training and evaluate in parallel, we use \texttt{Estimator} to distribute work across worker machines and parameter servers. There are currently two distribute strategies: \textit{replication} and the AdaNet-specific \textit{round-robin}. \texttt{Estimator} provides the default replication strategy, where workers replicate a copy of the full computation graph containing all candidates, and share variable updates through the parameter servers. In the second strategy, candidate subnetworks are placed on dedicated workers in a round-robin fashion. This is only possible because subnetworks can be trained independently from one another. Certain designated workers load every read-only subnetworks, and train only the ensemble parameters. The round-robin strategy reduces the load on the workers and parameter servers, speeds up training when subnetworks are large, and allows the system to scale linearly with the number of subnetworks and workers.

The system's distributed training and search execution is fault tolerant. If a worker or a cluster terminates due to preemption or an exception, it restores itself from a checkpoint to continue training with minimal loss in training time. 

In order to support the adaptive computation graph during distributed training, one worker is designated as chief, and is responsible for bookkeeping. Other workers idly loop until the chief writes the expanded checkpoint with an incremented iteration number.

\section{Applications and Results}%
\label{sec:results}
%

% \begin{figure}  
% \begin{center}  
% \includegraphics[width=8cm,keepaspectratio]{images/benchmark.pdf}
% \caption{\small \sl A comparison of algorithm performance on a set of production tabular datasets which include binary classification and regression tasks.}
% \end{center}
% \vspace{-0.3in}
% \label{fig:benchmark}
% \end{figure}  

\begin{figure}  
\begin{center}  
\begin{subfigure}[b]{0.45\textwidth}  
\includegraphics[width=6cm,keepaspectratio]{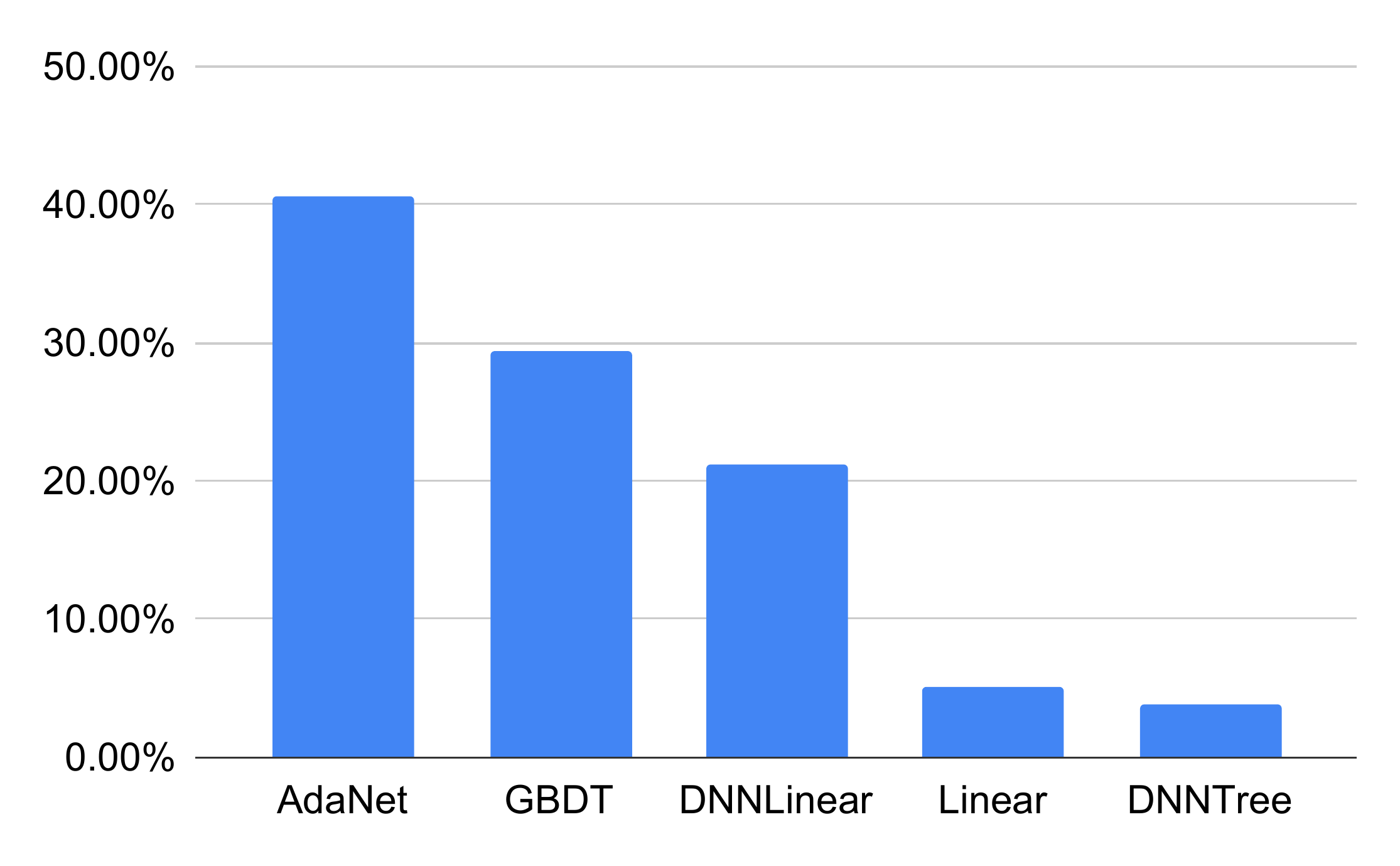}
% \label{fig:benchmark}
\caption{\small \sl Percentage of times an algorithm achieved the best performance on a set of production tabular datasets which include binary classification and regression tasks.}
\end{subfigure} 
\begin{subfigure}[b]{0.5\textwidth}
\includegraphics[width=8cm,keepaspectratio]{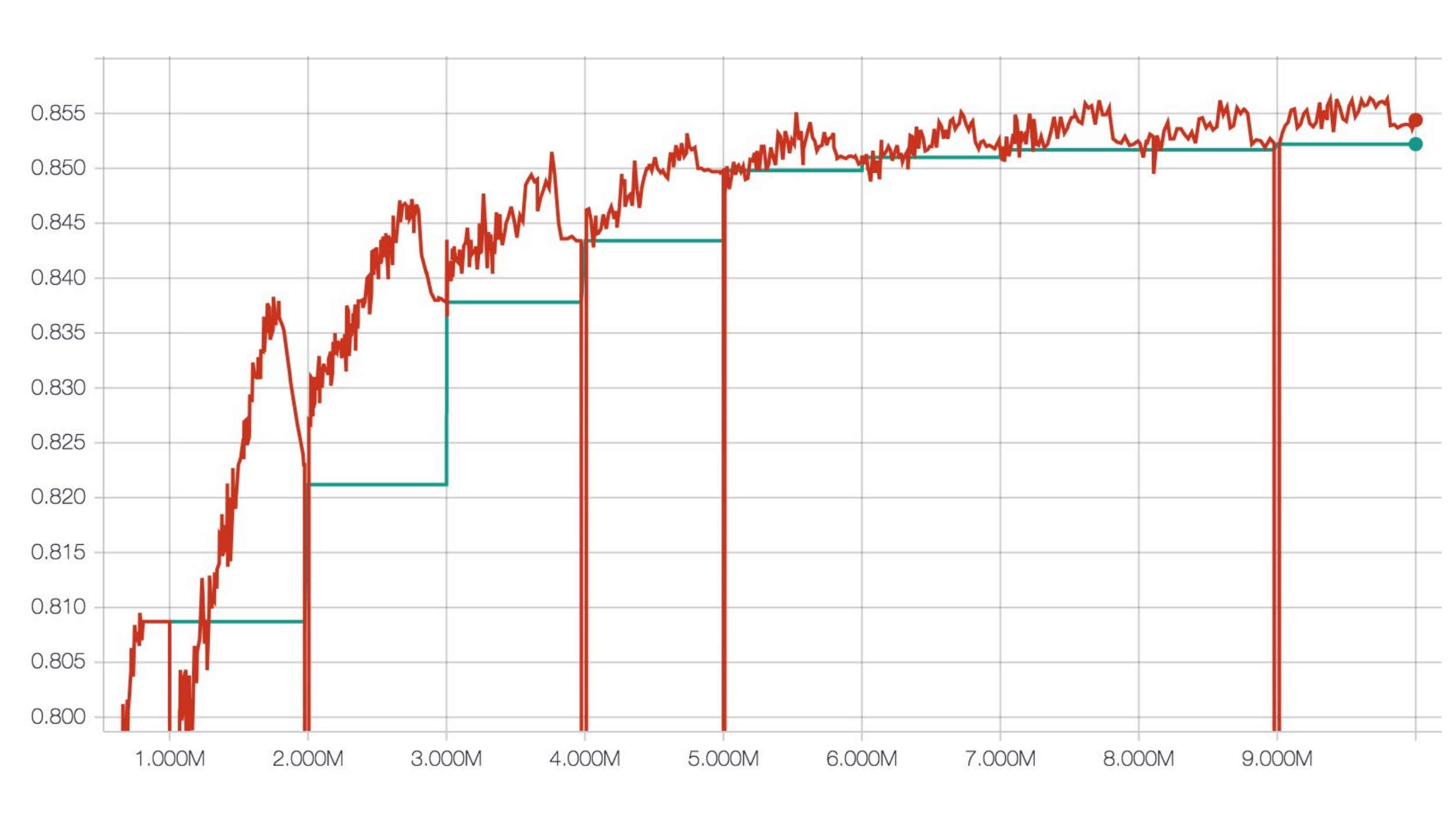}
\label{fig:search}
\vspace{-0.2in}
\caption{\small \sl Test accuracy per train step on CIFAR-100 using NASNet-A subnetworks. A new subnetwork begins training every million steps, and eventually improves the performance of the ensemble.}
\end{subfigure} 
\end{center}
\vspace{-0.2in}
\caption{\small \sl Examples of successes from applying AdaNet to different tasks.}
\label{applications}
\end{figure}

We battle-tested AdaNet through several engagements and launches with Google products. Each of the following applications applied our predefined \texttt{ComplexityRegularizedEnsembler} to learn scalar mixture weights that balances the trade-off between ensemble train loss and \textit{complexity} of the underlying subnetworks in order to obtain the learning guarantees from the algorithm from \cite{cortes17adanet}. See Appendix~\ref{sec:theory} for an overview of the theory behind the algorithm.

In one case, we integrated AdaNet into an internal platform designed for structured data using a search space composed of linear models, fully-connected neural networks that add a hidden layer at each iteration, and gradient-boosted decision trees. We benchmarked our search space versus other popular algorithms including wide-and-deep models, gradient-boosted trees, on over a hundred tabular datasets with thousands of datapoints. We also used bayesian optimization to tune each algorithm's hyperparameter set and trained each model for two hours using 10 CPU workers. Adanet produced the best model 40.56\% of the time, followed by gradient boosted trees 29.44\% (Figure~\ref{applications}).

In another case, the production model was already  an ensemble of 20 large and complex neural networks, built with custom ensembling infrastructure and trained on 200 CPU workers and 40 parameter servers. Using \texttt{AutoEnsembleEstimator} with the round-robin distributed strategy, we replaced the custom infrastructure with a simpler and more robust system, and enabled them to iterate more quickly by trying different search spaces.

Finally, in our own research, we applied AdaNet with convolutional subnetworks using the NASNet-A structure from \cite{zoph2017learning} to CIFAR-10 and CIFAR-100 to achieve 2.26\% and 14.58\% error rates respectively.  See Figure~\ref{applications} for a TensorBoard from a run training with 10 distributed V100 GPUs over 10 iterations of 1M steps each.

\section{Conclusion and Future Work}

% If you're interested in trying AdaNet for yourself, please check out our Github repo, and walk through the tutorial notebooks. We've included a few working examples using dense layers and convolutions to get you started. AdaNet is an ongoing research project, and we welcome contributions. We’re excited to see how AdaNet can help the research community.

AdaNet is a flexible and scalable framework for training, evaluating, and deploying ensembles of TensorFlow models (e.g. deep neural networks, trees, and linear models). It provides AutoML capabilities including automatic search over a space of candidate ensembles, supports CPU, GPU, and TPU hardware, and can scale from a single process to a cluster seamlessly using \texttt{tf.estimator.Estimator} infrastructure. The framework is flexible and can be extended to include a prior (i.e. fine-tuned production models) in its search space. It offers several out-of-the-box means of training and ensembling them (e.g. uniform average weighting, learning mixture weights). We open-sourced AdaNet to enable the AutoML community to leverage the large-scale computational offerings of industrial cloud providers just as we do within Google to accelerate research.

% Some immediate improvements can be made to the framework by plugging into state-of-the-art hyperparameter tuning to improve subnetwork quality. 
Future infrastructure work involves improving the ability to scale the number of candidates per iteration, and better handling the case when there are more candidates than machines. Future research involves providing better search spaces out-of-the-box for a wide range of common machine learning tasks, and developing new search algorithms for producing better ensembles more quickly, consistently,
 and in a principled manner.
% Acknowledgements should go at the end, before references and appendices
% \acks{We would like to acknowledge the contributions of our open-source contributors, and teams at Google whose discussions and feedback led to this design. We also thank Natalia Ponomareva and Dmitry Storcheus for their review of this manuscript.}

\vskip 0.2in
\bibliography{sample}

% Appendix goes to a new page

\newpage

%%%%%%%%%%%%%%%%%%%%%%%%%%%%%%%%%%%%%%%%%%%%%%%%%%%%%%%%%%%%%%%%%%%%%%%%%%

\appendix

\section{Complexity Regularized Ensembler Theory}%
\label{sec:theory}
Let $x$ be some input feature vector in the input space $x \in \cal X$, the ensemble consists of $l$ base-learner functions $\{h_k : h_k \in {\cal H}_k, k \in [l]\}$, where ${\cal H}_k$ is a class of functions so that $h_k: x \to \mathbb{R}^{n_k}$ where $n_k$ is the dimension of the output. The function $f$, which represents the ensemble that we are trying to learn, is a convex sum of the base-learner functions:
\begin{equation}
\label{eq:f}
f = \sum_{k=1}^l w_k h_k.
\end{equation}

Building an automatic ensemble of neural networks has two main challenges: (1) choosing the best subnetwork architectures; and (2) using the right number of subnetworks.
The AdaNet framework supports different ways to handle the learning of the weights $w_k$ and tackle these two challenges. A very robust baseline is obtained by having a uniform ensemble where the weights are $w_k = \nicefrac{1}{l}$. 
More advances techniques are related to complexity and generalization theory. While having numerous parameters is necessary for a neural network to obtain high expressivity power, they may not generalize to unseen data due to their greater complexity which could make the network memorize noise.
As proposed in~\citep{cortes17adanet}, the solution for the first problem is to define an upper bound on the Rademacher complexity of a DNN (see Section~\ref{sec:bounds}). The second challenge is related to the contribution of a DNN to the general ensemble (see Section~\ref{sec:balance_empirical_risk}).

%What are the best subnetwork architectures to consider? Is it best to reuse the same architectures or encourage diversity? While complex subnetworks with more parameters will tend to perform better on the training set, they may not generalize to unseen data due to their greater complexity. These challenges stem from evaluating model performance.
%Having numerous parameters is necessary for a neural network to obtain high expressivity power. Because of its great capability to capture any flexible data representation, deep neural networks have achieved great success in many applications.
%
%We could evaluate performance on a hold-out set split from the training set, but in doing so would reduce the number of examples one can use for training the neural network.

\subsection{Definitions and notation}

To simplify the presentation, we restrict our attention to the case of binary classification. All our results can be straightforwardly
extended to multi-class classification, by augmenting the number of output units and by using existing multi-class counterparts of the margin bounds.

Let $\cal X$ be the input space, and $\{-1, +1\}$ be the output space for the function we are trying to learn. The training set $S$ consists of $m$ labeled examples, which are drawn from the true distribution $\dataset$ over ${\cal X} \times \{-1, +1\}$. The ensemble consists of $l$ base-learner functions $\{h_k : h_k \in {\cal H}_k, k \in [l]\}$, where ${\cal H}_k$ is some class of functions that maps $\cal X$ to $\{-1, +1\}$. The function $f$, which represents the ensemble that we are trying to learn, is a convex sum of the base-learner functions:
\begin{align}
\label{eq:f2}
f = \sum_{k=1}^l w_k h_k, \quad\quad \text{s.t.} \quad \sum_{k=1}^l |w_k| \leq 1.
\end{align}

\subsection{Bounding the generalization error of ensembles}
\label{sec:bounds}

%To answer this question the framework leverages the theoretical properties of ensembles.

Our learning bounds are expressed in terms of the
Rademacher complexities of the hypothesis sets ${\cal H}_k$. The generalization error bound of an ensemble $f$ is given by the Deep Boost theorem\citep{cortes14deepboosting} as: 
\begin{equation}
\label{eq:constraint2}
R(f) \leq \hat{R}_{S, \rho}(f) + \frac{4}{\rho} \sum_{k = 1}^{l} \big | w  _k \big | \mathfrak{R}_m({\cal H}_k) + \widetilde O\left(\frac{1}{\rho} \sqrt{\frac{\log l}{m}}\right)
\end{equation}
where $\rho \in (0, 1]$, $R(f)$ is the generalization error $\E_{(x, y) \sim \dataset} [ \mathbbm{1}_{yf(x) \leq 0}]$, $\hat{R}_{S, \rho}(f)$ is 
the empirical margin error $\E_{(x, y) \in S}[\mathbbm{1}_{yf(x) \leq \rho}]$, and $\mathfrak{R}_m({\cal H}_k)$ is the Rademacher complexity of the function class ${\cal H}_k$ with $m$ examples.

Learning guarantees in Equation~\ref{eq:constraint2} are finer than those
that can be derived via a standard Rademacher complexity analysis~\citep{koltchinskii2002}. This is because it admits an explicit dependency on the mixture weights $w_k$ that allows a balanced contribution of subnetworks with very distinct complexities.

\subsection{Balancing empirical risk and generalization}
\label{sec:balance_empirical_risk}
AdaNet approach~\citep{cortes17adanet} seeks to find a function $f$ by optimizing the set of weights $\mathbf{w}$ in Equation~\ref{eq:f2} using the constraints of Equation~\ref{eq:constraint2}. The idea is to balance the trade-off between the ensemble’s performance on the training set and its ability to generalize to unseen data:
\begin{equation}
\label{eq:objective2}
F(\mathbf{w}) = \underbrace{\frac{1}{m} \sum_{i=1}^{m} \Phi \left (1 - y_i\sum_{j=1}^{N}w_jh_j \right )}_{\text{Empirical error}}
+
\underbrace{\sum_{j=1}^{N} \left (\lambda \cdot r(h_j) + \beta   \right )\left | w_j \right |}_{\text{Complexity penalty}}
\end{equation}
where $h_j$ is the $j$-th subnetwork, $w_j$ is the weight of the $j$-th subnetwork, $\Phi$ is the surrogate loss function, $r(h_j)$ is model $j$'s complexity, and $\lambda$ and $\beta$ are tunable hyperparameters. $\Phi$ is a non-increasing convex function (e.g., exponential function as in AdaBoost~\citep{adaboost}). Its purpose is to facilitate a convex surrogate of the empirical error.

By optimizing the weights of the ensemble $\mathbf{w}$ in Equation~\ref{eq:objective2} the model will include a candidate subnetwork only if it improves the ensemble’s training loss more than it affects its ability to generalize. This guarantees that the generalization error of the ensemble is bounded by its training error and complexity while simultaneously minimizing this bound.

The main benefit of optimizing the objective in Equation~\ref{eq:objective2} is to eliminate the need for a hold-out set for choosing which candidate subnetworks to add to the ensemble. This has the added benefit of enabling the use of more training data for training the subnetworks.

\subsection{Complexity and sensitivity measures}%
\label{sec:complexities}
The idea of complexity is to regularize the learning of mixture weights of each subnetwork based on their sensitivity to training data as a proxy for the potential generalization error on test data.

In addition to the Rademacher complexity used in Section~\ref{sec:bounds} and the upper bounds presented in~\citep{cortes17adanet} we have introduced two new sensitivity measures that work as proxy functions of the Rademacher complexity: (1) the norm of the variance at the final layer across data points for each batch; and (2) the matrix norm of the Jacobian of the logits layer with respect to the input layer.

The motivation for introducing additional complexity measures comes from our experimental observations on the behaviour of DNNs over many datasets.
While there is a general relationship between Rademacher complexity and the size of a subnetwork, it can be loose because the Rademacher complexity also depends on the shape of the architecture. %Hence, it is a natural evolution to some of the common problems with Rademacher upper bounds when dealing with very deep neural networks.
This was supported by our preliminary experimental results in~\citep{cortes17adanet} where it was shown that the variance can perform almost as well in regularizing the mixture weights.
The idea has to do with that of variance-bias trade-off~\citep{variance_bias}. With similar training losses one learner with a higher variance could be a sign of fitting to random noise in each training batch. \cite{novak2018sensitivity} explored the norm of the input-output Jacobian of the network and found it to correlate well with generalization on simplified fully-connected networks.
The hypothesis is that both measures can fare as least as good as our Rademacher-based bound on convolutional networks as well as deeper fully connected networks.

Variance of subnetwork $h_k$ is calculated as,
\begin{align*}
\sigma^2(h_k) &= \frac{1}{m}\sum_{j=1}^m (h_k(x_j) - \overline{h}_k) \\
\overline{h}_k &= \frac{1}{m} \sum_{j=1}^m h_k(x_j),
\end{align*}
where $h_k(x_j)$ is the output of subnetwork $k$ for the $j$-th input example.

Similar patterns were seen for the Jacobian but at a lower correlation. As inspired by~\citep{novak2018sensitivity} we introduce the Frobenius norm of the Jacobian matrix at the softmax layer. Accordingly, this is obtained by computing the matrix of point-wise derivatives of the number of outputs activations and also computes the Euclidean matrix norm before averaging over the training points.

\end{document}